# Combined Haar-Hilbert and Log-Gabor Based Iris Encoders

V.E. Balas, I.M. Motoc, A. Barbulescu

**Abstract** This chapter shows that combining Haar-Hilbert and Log-Gabor improves iris recognition performance leading to a less ambiguous biometric decision landscape in which the overlap between the experimental intra- and inter-class score distributions diminishes or even vanishes. Haar-Hilbert, Log-Gabor and combined Haar-Hilbert and Log-Gabor encoders are tested here both for single and dual iris approach. The experimental results confirm that the best performance is obtained for the dual iris approach when the iris code is generated using the combined Haar-Hilbert and Log-Gabor encoder, and when the matching score fuses the information from both Haar-Hilbert and Log-Gabor channels of the combined encoder.

## 1. Introduction

In 1970s, Flom and Safir [11], two American ophthalmologists noticed that the iris texture differs from one person to another and later asked Daugman to develop a system for identifying persons using their iris. The system patented by Daugman in 1994 [4] and based on a 2-dimensional Gabor filter was the first fully functional iris recognition system. In the same period, Wildes et al [53] proposed a different iris recognition system.

Compared to 1990s, iris recognition is nowadays a relatively popular research topic, many new segmentation, encoding or matching methods being proposed in the last two decades as original solutions produced by well-established research teams from Bath University ([29], [47]), CASIA ([23]-[25], [54]-[57]), NIST ([13], [31]), Notre-Dame University ([1], [18], [22], [33]), Kent University ([46], [51], ) or by individual researchers around the world: L. Masek - [28], C. Tisee [59], S. Yang [62], S. Yoon [63], S. Ziauddin [66] and many others. Approaches relying on soft computing techniques, logical formalism and neuro-evolutionary architectures for iris recognition systems were recently proposed by N. Popescu-Bodorin, V.E. Balas and I.M. Motoc in [36] and [39]-[43]. Some airports and seaports around the world (Arab Emirates, for example) decided to increase their security and they achieve that using iris recognition system, whereas in

Valentina E. Balas, IEEE Senior Member, Associate Professor,
 Faculty of Engineering, *Aurel Vlaicu* University of Arad, Arad, Romania,
 email: balas@drbalas.ro
Iulia M. Motoc, IEEE Student Member,
 Artificial Intelligence and Computational Logic Laboratory, Department of Mathematics
 and Computer Science, *Spiru Haret* University, Bucharest, Romania,
 email: motoc@irisbiometrics.org
Alina Barbulescu, Associate Professor,
 Faculty of Mathematics and Computers Science, *Ovidius* University of Constanta, Romania,
 email: abarbulescu@univ-ovidius.ro

Afghanistan, authorities plan to scan the irides of the entire population. On the other hand, the global market of biometric devices and technologies continues to expand - some of them recently being tested by NIST [13], [31].

The independent evaluation undertaken by NIST for the most popular iris recognition technologies available in 2007-2009 established a characterization of what is the present state of the art in iris recognition, and established a framework for future progress and a level of understanding the iris recognition theory and practice.

When we say "iris recognition" we understand that an artificial agent extracts and matches some iris codes in order to produce a biometric decision (accept/reject) accordingly to a computed similarity score assumed to encode in some degree the actual similarity between the iris images currently being compared. This is a human-made agent and therefore we might be tempted to think that it will behave like a human. However, things are far from being this way: on the one hand IREX report shows that the present *state of the art* in iris recognition is still grounded to a statistical decisional landscape in which the biometric decision is bimodal (the two distribution of scores overlap each other creating a confusion zone), and on the other hand, in [41] and [42] it is shown in what conditions the statistically confused score distributions could generate a binary consistent artificial understanding of iris recognition.

## 2. Perceiving iris recognition through Turing tests

The idea of undertaking Turing tests [60] for iris recognition originated in [44], some results of such tests being published already in [40] and [42]. The importance of these tests resides in the fact that they certify the distance between the present state of the art statistical (bimodal) iris recognition and a prototype recognition function identified while interrogating the human agent during the Turing test. Such a test leads to an inevitable comparison between how a human perceives and performs the act of iris recognition (Fig.1.a from [42]) and how an artificial agent (for example, a state of the art iris recognition system, which takes statistically motivated biometric decisions - like those tested in [13]) operates the iris recognition task.

During a Turing test of iris recognition, it's easy for a human agent to see if between two irides is or isn't a difference, hence, as it is said in [42], the geometry [64] of his decisions is a crisp one (Fig. 1.a, [42]) and consists of one collection of crisp points (0 and 1) and a histogram that shows how many times the human agent recognized two irides as being similar (when the decision is given with an unitary score) or as being different (when the decision is encoded as a null score).

However, despite the fact that for a software agent the recognition is a much-complicated task, the recognition results in this case are not necessarily as accurate and correct as those given by the human agent. The human agent sees the genuine and imposter comparisons as two crisp and disjoint concepts, but for the artificial agent (for an iris recognition system practicing the present bimodal statistical recognition) these concepts are fuzzy and statistically confused along the zone where the imposter and genuine score distributions overlap each other. This overlapping defines here the confusion zone.

A human needs only two values (0 and 1) to encode the meaning of the two different, complementary and mutually exclusive concepts ('genuine' and 'imposter' comparisons) whereas the machine will encode the similarity between the two irides using certain methods of computing a similarity score belonging in [0, 1] interval. Hence, the concepts 'genuine' and 'imposter' are crisp in human perception / understanding and they are fuzzy in the artificial perception of an artificial software agent.

Nevertheless, despite the way in which the artificial and human agents perceive them, what really matters is how these two concepts really are: they are distinct (disjoint), complementary and mutually exclusive.

The fact that in the artificial perception of some iris recognition system the two concepts are seen as being insufficiently distinct (the genuine and the imposter scores define two fuzzy intervals which are different in meaning - see the concept m-precisiation, Zadeh [65], but they overlap each other), and consequently not quite complementary and not quite mutually exclusive, is not a reason to believe that the iris recognition and those two concepts are indeed fuzzy in their nature. This fact proves only how much room for improvement exists between two paradigms of iris recognition, namely between the statistical / bimodal iris recognition ([1]-[7], [13]), on the one hand, and on the other hand, the intelligent and logical recognition certified as being possible by the Turing test and recently studied in [44] and also in [40]-[43].

## 3. EER vs. f-EER

It has been shown in [42] that the decisions given by the artificial agent during a Turing test of iris recognition defines an f-geometry (Zadeh, [64]) in which the inter- and intra-class score distributions could or could not overlap each other.

On the other hand, as Daugman said in [7], the recognition errors are caused by the intersection of the genuine and imposter distributions (the system is more efficient if the error is insignificant, almost zero, meaning that greater the distance between the two distribution of scores, better the iris recognition system is).

To express the fuzzification between inter- and intra-class score distribution, a crisp concept known as Equal Error Rate (EER) is usually used, but as it is said in [44] and also in [38]-[41], the existence of such a crisp point was not experimentally confirmed. In our experience also, the theoretical concept of EER corresponds to a collection of possible EER points that are varying from one recognition test to another. This collection of varying EER points can be seen as a fuzzy EER interval (denoted f-EER in [42]), which, as it is said in [43], means "*a collection of recognition thresholds for which is very hard (or simply impossible) to say for sure if they are recognition scores rather than rejection scores or vice versa*".

The f-EER interval is the f-geometry (Zadeh, [64]) corresponding to the crisp and theoretical concept of EER point. This happens every time when the genuine and imposter score distributions overlap each other along a confusion zone which causes an eventual binary logical model of iris recognition to collapse (to become logically inconsistent) because people which are not enrolled in the system may have the possibility to pass as they would be a genuine match for an enrolled person.

One way to eliminate the inconsistency ([42], [44]) is adding a third fuzzy set in-between the fuzzy intervals containing the genuine and imposter scores, namely the f-EER interval, which is seen like defining a safety band that enable the biometric system to keep the inter- and intra-class score distributions disjoint.

More exactly, f-EER corresponds to a third logical state "u" (uncertain / unknown) different from 0 and 1. The similarity scores belonging in this region will label as unenrollable or undecidable those pairs of irides (or pairs of iris codes) for which the artificial agent (the iris recognition system) could not say exactly if they are genuine or imposter pairs, indeed.

## 4. Iris segmentation, encoding and matching

This section describes three iris texture encoders: Log-Gabor, Haar-Hilbert, and the combination between them. All of them will be tested further in this paper in the single eye enrollment scenario and also in the dual iris approach (proposed in [45] and [46]). One of our goals here is to test the combination between the dual iris approach and the fusion of the two classifiers based on Log-Gabor and Haar Hilbert encoders, respectively.

Iris segmentation is practiced here using the Circular Fuzzy Iris Segmentation procedure (CFIS2) proposed in [38] and available for download in [39]. For any input eye image from the test database, the result of this segmentation procedure is a concentric circular ring (delimited by a circular approximation of the pupil at the interior and by a circular approximation of the limbic boundary at the exterior) or equivalently, a rectangular polar unwrapped iris segment whose lines are iris circles unfolded in the angular direction.

Relative to the iris segment extracted with CFIS2, an additional correction is applied here for the pupil segmentation by eliminating from the unwrapped iris segment the lines situated near the pupil and which accidentally contain a certain amount of pixels from the pupil. All the unwrapped iris segments are further normalized at the dimension of 256x16.

### 4.1. The Log-Gabor iris texture encoder

The Gabor filters were introduced in 1940 [12] as signal processing techniques, then studied by Helstrom [16], Montgomery [30], and Rihaczek [49]. They later came into the attention of some researchers focused on understanding the way in which the cells of visual cortex within the mammal's brain sustain the complex process of vision and on finding a computational model for the human vision. Marčelja ([27], 1980) and Field ([10], 1987) relied on them to describe the functionality of receptive fields of the visual cortex, and Daugman ([3], 1988) used them initially as image compression tools and later for phase-based encoding of the iris texture ([4]-[8], 1994-2007).

When it came to choose between 1D or 2D Log-Gabor encoding, we selected the former because, as in [44], we also found that as long as the equations describing the iris movement in the radial direction will remain unknown the attempt of matching irides in the radial direction will also remain an insurmountable source of errors, especially inconsistent (insufficiently motivated) False Reject errors. Consequently, the Log-Gabor filter used in this paper is a single-scale, fast, one-dimensional variant of the encoder used in [28], it encodes the phase of iris texture only in the angular direction and has the following form:

$$G(f) = \exp[-0.5\log^2(f/f_0)/\log^2(\sigma/f_0)],$$

where $f_0$ is the center frequency, $\sigma$ is the bandwidth of the filter.

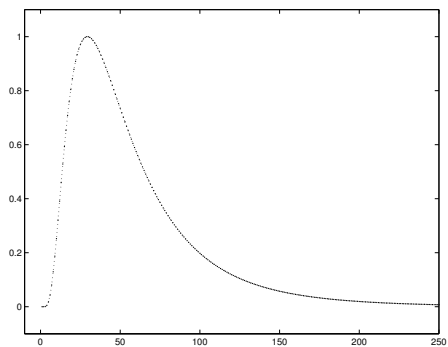

**Fig. 1.** Displaying the Log-Gabor filter in the frequency domain for an iris line of 512 pixels: the highest 256 FFT frequencies and the DC component are neglected all together (see the left-side and the right-side of the graphic), whereas the other components (those corresponding to the remaining frequency bands which are not neglected) are rescaled through an affine combination whose components draw a Gaussian when represented against a logarithmically scaled abscise.

The same encoder is used in [38], [39] and [44] which, together with [29] and the IREX Report [13], are taken as references for the iris recognition results further presented in this paper and obtained for the same iris database [52].

The Matlab implementation of the above one-dimensional single-scale Log-Gabor encoder uses the Fast Fourier Transform (FFT) and its inverse and is available for download within the toolbox [39]. It encodes the polar unwrapped and normalized iris segment line by line, in the angular direction.

The Log-Gabor encoder presented above compress the lines of the normalized iris segment in the frequency domain in a lossy manner, by neglecting the highest 128 FFT frequencies together with the DC component. It also enhances those frequency bands that store the discriminative information on which intra-class matching and inter-class rejection are both based on.

## 4.2. The Haar-Hilbert iris texture encoder

The Haar-Hilbert encoder was introduced in [44] and [38]. It encodes the iris texture as a binary matrix and has two operations:

–   The first one consists in a single-level 2-dimensional Discrete Haar Wavelet decomposition applied on the normalized iris segment in order to smooth it by removing a 2-dimensional noise signal involuntarily and artificially introduced there during the image acquisition and during the preprocessing stages which took place while transforming the initial iris image into the normalized rectangular iris segment. The existence of this noise was for the first time assumed, experimentally verified and documented in [44]. Our iris recognition results that follow to be presented here also confirm this hypothesis. This denoising operation is an operation at the global scale of the normalized iris segment and produces a denoised iris segment of dimension 128x8 in our case.

–   In the second step, the Hilbert transform is applied to the denoised iris segment block-wise (locally) in the angular direction. The result is a block-wise strong analytic signal [36] whose phase is further encoded as a binary iris code. Generally, in this paper, each time when a block-wise operation is performed, the dimension of the processing block is mentioned within the table where the experimental data are reported.

Summarizing, the Haar-Hilbert filter consists in a global denoising of the normalized iris segment followed by a local (block-wise computed) binary phase-based very lossy compression of the strong analytic signal generated locally by block-wise computing the Hilbert Transform of the denoised iris segment.

One of the most intuitive ways to introduce the Hilbert Transform and to take a meaningful view over the related topics was described by M. Johansson in [17] where he drew an imaginary path from the exponential form of the complex numbers (Euler, [9]):

$$e^{jz} = \cos(z) + j \cdot \sin(z),$$

to the complex notation of harmonic waves (generalized Moivre's formula written for the exponential form of the complex numbers):

$$e^{j\omega t} = \cos(\omega t) + j \cdot \sin(\omega t),$$

and to the basic property of the Hilbert Transform - namely that relative to the input signal, it shifts the phase of all frequency components by $\pi/2$ radians, property proved by Hilbert as a consequence of the fact that $\sin(\omega t)$ is the Hilbert Transform of $\cos(\omega t)$. For a given initial signal x it follows then very naturally the introduction of Gabor analytic signal y, defined in [12] as:

$$y = x + j \cdot H(x).$$

where the Hilbert Transform of a continuous time-domain signal f is defined as:

$$H(f(t)) = \frac{1}{\pi} P \int_{-\infty}^{+\infty} \frac{f(t)}{t - \tau} d\tau,$$

whenever the integral exists.

According to [36], the reason for which the Hilbert Transform is suitable to be used in iris recognition is that the energy of a signal is an invariant of the Hilbert Transform. Other properties of Hilbert Transform can be found in [19] whereas Fig. 1 from [35] shows an intuitive depiction of the binary phase encoding based on the Hilbert Transform, encoding defined by the following relation:

$$BIC = logical(\ phase(y) > 0\ ),$$

which defines the Gabor Analytic Iris Texture Encoder [35] and establishes that the k-th bit of the binary iris code BIC has a value of 1 if and only if the corresponding component $y(k)$ of the Gabor analytic signal y has a positive phase.

## 4.3. Combined Haar-Hilbert & Log-Gabor encoder. Classifier fusion strategy

The fact that the Log-Gabor filter encodes the normalized iris segment in the angular direction, line by line on their full length, means that the phase features detected with the Log-Gabor encoder are relevant at the scale of the iris circles concentric to the pupil, or in other words, they are meaningful at the global scale of the circular iris segment. On the contrary, the phase features encoded by the Haar-Hilbert filter are computed block-wise, meaning that they are locally relevant in the first place, instead of being meaningful at the global scale of the circular iris segment.

However, each encoder defines a fuzzy binary classifier (or a binary-modal classifier – we could say) which organizes the set of all iris code pairs within the test database into two (modal) classes, namely the genuine and the imposter pairs respectively, classes that expose to each other statistically confused sparse fuzzy boundaries. From this perspective, to improve the quality of iris recognition should mean to move these boundaries away from each other and toward their own classes. On the other hand, the sparsity of these fuzzy boundaries tells us that in an exhaustive test of iris recognition made by following the single eye enrollment scenario, obtaining a similarity score situated in the confusion zone is a rare event. Hence, it makes sense asking how much correlation it is between these rare events produced on the two different processing channels of the combined Haar-Hilbert & Log Gabor (HH&LG) classifier illustrated in Fig. 2, how much correlation it is between two rare events occurring on two different channels of information from which one encodes global features and the other encodes localized features within the phase of the iris texture. The experimental work underlying this paper shown us there is a certain degree of independence between these rare events occurring on the two processing channels, not very high, but high enough to ensure an improvement of iris recognition when the two classifiers are fused.

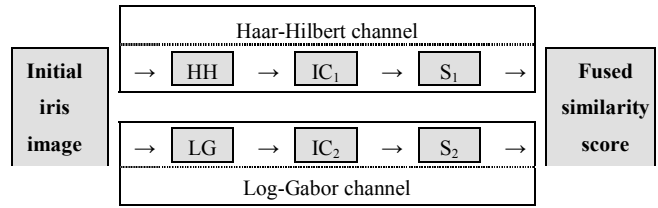

**Fig. 2.** Iris recognition based on classifier fusion and single eye enrollment scenario

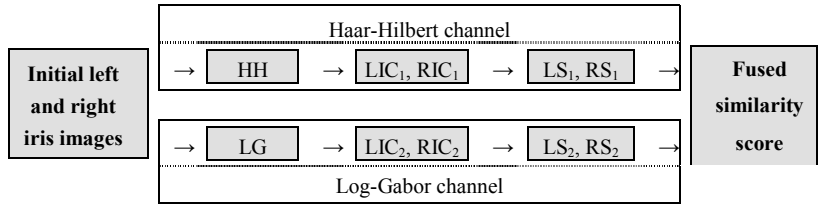

**Fig. 3.** Iris recognition based on classifier fusion and dual iris approach

The combined Haar-Hilbert & Log-Gabor classifier is illustrated in Fig. 2: an iris image I is acquired for the current candidate C and two candidate iris codes $IC_1$ and $IC_2$ are generated from the image I using the Haar-Hilbert and Log-Gabor encoders. Then the candidate iris codes $IC_1$ and $IC_2$ are matched against two binary templates stored under a certain claimed identity E using the Hamming distance and two similarity scores $S_1$ and $S_2$ are computed as the results of these comparisons. The membership degree of the current candidate C to the claimed identity E is further computed as a fused similarity score:

$$S = (S_1 * S_2)^{1/2}.$$

In the dual iris approach (Fig. 3), four binary iris codes ($LIC_1$, $LIC_2$, $RIC_1$, $RIC_2$) and four similarity scores are generated as described above for the two eyes of the current candidate C: $LS_1$, $LS_2$ (for the left eye of the candidate), $RS_1$ and $RS_2$ (for the right eye). The membership degree of the current candidate C to the claimed identity E is then computed as:

$$S = (LS_1*LS_2*RS_1*RS_2)^{1/4}.$$

## 5. Performance criteria

The results of the experimental iris recognition tests that follow to be presented here will be compared using classical performance criteria such as the decidability index, Fisher's ratio, the False Accept Rates (FAR), the False Reject Rates (FRR) and the Equal Error Rate (EER), but also using non-standard criteria recently introduced in [44], [42] and [43] such as the Pessimistic Odds of False Accepts (POFA), Pessimistic Odds of False Rejects (POFR), Pessimistic Odds of Equal Error (POEE), the overlap between the imposter and genuine experimental distribution of scores, the safety interval ([42], [44]), the compatibility with a Fuzzy 3-Valent Disambiguated Model (F3VDM, [42], [44]) of iris recognition and the type of iris recognition theory ([43], [44]) exhibited by the recognition system during the tests.

### 5.1. Decidability index and Fisher's ratio

In iris recognition, Daugman ([4], [5]) introduced the decidability index in order to express the degree of separation (the statistical bimodal separation) between the inter- and intra-class score distribution as being the number d' computed as follows:

$$d' = |\mu_I - \mu_G| / [(\sigma_I^2 + \sigma_G^2)/2]^{1/2}.$$

where $\mu_I$, $\mu_G$, $\sigma_I$ and $\sigma_G$ are the means and the standard deviations computed for the imposter and genuine experimental score distributions. Nevertheless, we traced the use of such formulas (with little variations, but with the same meaning) back to 1954 in the writings of Peterson, Birdsall & Fox - [32] and Tanner & Swets - [58].

Within the same family of separation measures is also the Fisher's ratio whose use in iris recognition was suggested by Wildes in [61]:

$$FR = (\mu_I - \mu_G)^2 / (v_I + v_G),$$

where $\mu_I$ and $\mu_G$ are defined above, whereas $v_I$ and $v_G$ are the variances of the two experimental distributions of imposter and genuine scores, respectively. The decidability index d' and the Fisher's ratio FR are both estimated (optimistically) from data, their relevance being based on the hypothesis that the volumes of experimental imposter and genuine data have exceeded already some (a priori unknown) critical values above which the statistics of the two classes of scores become stationary. Since the test database used here is very small ([52], 1000 images) when compared to the world population, we have not found reasons to assume that the above hypothesis is satisfied, and therefore, as a precaution, we will prefer here to make a distinction between the data objectively measured during our iris recognition tests and the statistical measures optimistically or pessimistically estimated based on the actual experimental data. This explains the distinction that we make here between the actual *Rates* measured for the numerical results of our tests and the optimistically or pessimistically estimated *Odds* that some event to occur or not in the future exploitation of an iris recognition system that could hypothetically prolong a given recognition test undertaken here. The same distinction between *Rates* and *Odds* was practiced also in [38] and [44].

### 5.2. FAR, FRR, EER, OFA, OFR, OEE, POFA, POFR and POEE

For any given recognition threshold t, the False Accept Rate - FAR(t) - is defined here as in [36], as being the experimentally determined "*ratio between the number of imposter scores exceeding the threshold and the total number of imposter scores*", i.e. the cumulative of the actual experimentally determined imposter probability density function form the threshold t to the maximum imposter similarity score.

By analogy, for any given recognition threshold t, the False Reject Rate –FRR - is the experimentally determined as the "*ratio between the number of genuine scores not exceeding the threshold and the total number of genuine scores*" [36], i.e. the cumulative of the actual experimentally determined genuine probability density function form the minimum genuine similarity score to that threshold t.

The theoretical concept of EER point is then defined by the common value of the FAR and FRR curves at the threshold $t_{EER}$ where they equal each other. If the experimentally determined genuine and imposter probability density functions (pdf-s) are overlapping each other then the EER value is strictly positive.

Still, it could happen that the experimentally determined pdf-s are not overlapping each other. In this case, the EER value is null and it makes sense trying to predict an EER value for future exploitation of the system in terms of Odds of Equal Error (OEE) and Pessimistic Odds of Equal Error (POEE). Unlike the FAR and FRR, the *Odds of False Accept* (OFA), the *Odds of False Reject* (OFR) [36], the *Pessimistic Odds of False Accept* (POFA, [38]) and the *Pessimistic Odds of False Reject* (POFR) [44] are estimated from data.

The OFA and OFR are optimistically estimated from data by fitting the actual score distributions with theoretical ones, determined by their means and their standard deviations. However, these theoretical pdf-s are not necessary pessimistic envelopes (see such pessimistic envelopes in Fig. 7.a, Fig. 7.b and Fig. 7.c) for the actual experimental pdf-s and this is what makes them optimistic approximations of the actual pdf-s.

For any given recognition threshold t, OFA(t) is defined [36] as being:

$$\mathrm{OFA}(t) = \int_t^1 I_{pdf}(\tau)d\tau \, ,$$

i.e. the cumulative of the theoretical optimistically estimated imposter pdf ($I_{pdf}$) on the interval [t, 1], whereas OFR(t) is defined [36] as:

$$\mathrm{OFR}(t) = \int_0^t G_{pdf}(\tau)d\tau \, ,$$

i.e. the cumulative of the theoretical optimistically estimated genuine pdf ($G_{pdf}$) on the interval [0, t].

By analogy with EER, the Odds of Equal Error (OEE) are defined by the common value of the curves OFA and OFR at the threshold $t_{OEE}$ where they equal each other. Unlike EER value, which may be null sometimes, in the paradigm of bimodal iris recognition ([4], [5], [7], [36], [38]) the OEE value is always strictly positive (even when the EER value is null).

The pessimistic estimations of the actual genuine and imposter pdf-s (denoted $PI_{pdf}$ and $PG_{pdf}$) may be obtained from these theoretical (optimistically determined) pdf-s described above ($I_{pdf}$ and $G_{pdf}$) by scaling up their standard deviations with a certain (supra-unitary) factor (i.e. by accepting the pessimistic hypothesis that, over the time, the intraclass variability - encoded through the standard deviations of each class - would or could increase) or by slightly increasing/decreasing the mean of imposter/genuine distribution with a certain additive positive/negative shift [38], respectively (i.e. by accepting the pessimistic hypothesis that, over the time, the intraclass/genuine score distribution would slightly slide to the right/left toward the distribution of genuine/imposter similarity scores) or by combining the sliding of the genuine and imposter similarity score distributions toward each other with a growth of their variability. For any given recognition threshold t, POFA(t), POFR (t) ([44]) and Pessimistic Odds of Equal Error (POEE) defined by analogy with OFA(t), OFR(t) and OEE:

$$\mathrm{POFA}(t) = \int_t^1 PI_{pdf}(\tau)d\tau \, ,$$

$$\mathrm{POFR}(t) = \int_0^t PG_{pdf}(\tau)d\tau \, .$$

Evidently, the POEE will be always greater (i.e. more pessimistic) than the OEE. As a precaution, such pessimistic evaluation measures will be used further in this paper (see Table 1 and Table 2).

### 5.3. The compatibility with a Fuzzy 3-Valent Disambiguated Model

We recall that when the statistical / bimodal decisional model (introduced by Daugman, [4]-[7]) is implemented and practiced on an iris recognition system, the two concepts 'genuine' and 'imposter' are artificially perceived in the system as being not quite mutually exclusive despite that they are actually distinct, mutually exclusive and even complementary concepts, fact which can be verified during a Turing test of iris recognition (see Fig. 1.a from [42]). The confusion zone is defined by the minimum genuine and maximum imposter similarity scores whenever the former is smaller than the later. Let us call the *imposter interval* as being the interval determined by the extreme imposter values experimentally determined during an exhaustive test of iris recognition, and the *genuine interval* defined by analogy.

The confusion zone is the intersection of these two intervals whenever they overlap each other. In a favorable scenario, the genuine and the imposter interval are disjoint and the confusion zone is undefined. Hence, in these cases it makes sense to talk about the confusion zone in terms of *Odds*. The only problem is that the theoretical OFA, OFR, POFA and POFR have positive values anywhere in (0, 1) interval. It happens this way because, for example, as it is said in [44] there is no theorem to guarantee that all imposter scores belong naturally to a certain interval centered in 0.5. Hence, for the moment, the existence of such a crisp right boundary (a crisp majorant) for all imposter similarity scores is inevitably assumed as a pure hypothesis, based on the experimental data, which implicitly means to accept that the experimentally determined imposter pdf should possess a vertical asymptote [44] on its right side.

However, the existence of a vertical asymptotic behavior at the left side of the experimentally determined pdf of the genuine similarity scores is any but possible because, as it is exemplified in [41] and [44], the index of the genuine comparisons may be accidentally corrupted by comparing a wrong segmented iris segment taken from an image of a given eye to an iris segment correctly extracted from other image of the same eye, or in general, by comparing two very different hypostases of the same iris captured in very different acquisition conditions (for example, one with a very contracted pupil and the other slightly rotated and showing a very dilated pupil of the same eye – in this case the two images illustrate two configurations of the same physical iris, configurations which are so different that their matching is indeed impossible). The influence of pupil dilation on the iris biometric performance is also documented in [18].

Let us consider now an exhaustive iris recognition test in which the experimentally determined imposter and genuine intervals are statistically confused on their tails, or in other words, the separation between the two classes of scores is fuzzy. In this case, the size of the overlap is defined as the length of the interval on which the two classes of scores are overlapping each other, i.e. the difference between the maximum imposter score (MIS) and the minimum genuine score (mGS):

$$O_1 = MIS - mGS.$$

| I | [a, 1] | **Genuine Pairs** | Imposed Odds of False Accept: **POFA(a)** True Accept Safety: **1-POFA(a)** | | |
|---|---|---|---|---|---|
| **O** | (r, a) | **Artificially Undecidable Pairs** | Genuine Discomfort Rate < POFR(a); Imposter Discomfort Rate < POFA(r); ----------------------------------------(+)- Total Discomfort Rate < POFR(a)+ POFA(r); | INCERTITUDE | DISCOMFORT / SECURITY |
| **D** | [0, r] | **Imposter pairs** | Imposed Odds of False Reject: **POFR(r)** False Reject Safety: **1-POFR(r).** | | |

**Fig. 4.** A Fuzzy 3-Valent Disambiguated Model obtained by imposing the following two security restrictions: $r = POFR^{-1}(v_1)$ and $a = POFA^{-1}(v_2)$, where $v_1$ and $v_2$ are imposed values.

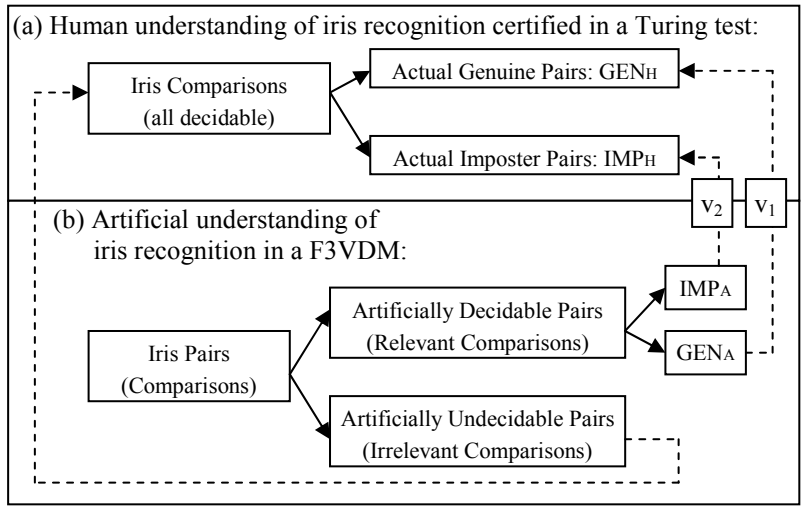

**Fig. 5.** (a) Crisp binary human understanding of the input space certified in a Turing test of iris recognition undertaken for the test database [52]. (b) Hierarchical binary classification of the input space associated to a Fuzzy 3-Valent Disambiguated Model of iris recognition when practicing the iris recognition within the limits of (logically) Consistent Biometry [40].

The results of an iris recognition test are better if the overlap size is smaller. Negative values of this parameter are desirable meaning that the two experimentally determined distributions of scores are separated by a safety interval whose width, as it is said in [44], negotiates between system security and user comfort. Regardless the overlap size, the similarity scores belonging in the safety interval are considered inconclusive for taking a biometric decision. This is the main feature of a Fuzzy 3-Valent Disambiguated Model (F3VDM): the fuzzy separation between the two classes of similarity scores is enforced to become a crisp one by introducing the safety interval in-between them.

The Fuzzy 3-Valent Disambiguated Models (F3VDM) of iris recognition are proposed in [44] and [41] as solutions to the following type of problems: given the results of an exhaustive iris recognition test, find a partition $\{[0, r], (r, a), [a, 1]\}$ of $[0, 1]$ (as in Fig. 4) satisfying an imposed functioning regime specified in terms of system security or in terms of user (dis)comfort.

A F3VDM reveals the contrast between the human and artificial understanding of iris recognition where the later is achieved by an artificial hardware-software agent (an artificial iris recognition system): as certified by the Turing tests of iris recognition (see Fig. 1.a in [42]) and as illustrated in Fig. 5.a from above, the human understanding classifies the pairs of iris codes in two distinct, mutually exclusive and complementary classes, namely the set of all imposter pairs $IMP_H$ and the set of all genuine pairs $GEN_H$, whereas in a F3VDM, the artificially perceived concepts $IMP_A$ and $GEN_A$ are mutually exclusive but no longer complementary. In other words, for any artificial iris recognition system, practicing the iris recognition within the limits of (logically) Consistent Biometry [40] means to achieve a hierarchical binary classification of the input space (the space of all iris code pairs) in four classes: undecidable/unenrollable pairs, decidable/enrollable pairs, decidable-imposter pairs and decidable-genuine pairs, as illustrated in Fig. 5.b from above.

Fig. 5.b describes the behavior of an artificial unsupervised iris recognition system. Its artificial understanding is described below as a *Qualitative Sugeno Model* [44] defined by the following three fuzzy if-then (Sugeno, [53]) rules written in Cognitive Dialect [37], as in [44]:

i)   (!:) $\{\{(!:)[ t \rightarrow (C \in IMP_A) ]\} \leftrightarrow \{(!:)[ t \rightarrow ((d \circ S)(C) = D) ]\}\}$,

ii)  (!:) $\{\{(!:)[ t \rightarrow (C \in UND_A) ]\} \leftrightarrow \{(!:)[ t \rightarrow ((d \circ S)(C) = O) ]\}\}$,

iii) (!:) $\{\{(!:)[ t \rightarrow (C \in GEN_A) ]\} \leftrightarrow \{(!:)[ t \rightarrow ((d \circ S)(C) = I) ]\}\}$,

where:
-   the values D, I and O encode Different irides (a decidable-imposter pair), Identical irides (a decidable-genuine pair) and Otherwise (an undecidable pair), respectively;
-   C denotes the current pair of irides;
-   S(C) is the similarity scores computed for the pair C;
-   d is a defuzzification / decision function defined by the following three relations:

$d(S(C)) = I$ if and only if $S(C) \in [a, 1]$;

$d(S(C)) = D$ if and only if $S(C) \in [0, r]$;

$d(S(C)) = O$ if and only if $S(C) \in (r, a)$;

- $UND_A$ is the set of those iris pairs which are undecidable in the artificial understanding of the iris recognition system;

Fig. 5.a illustrates the human understanding of iris recognition (which is crisp and binary) certified through a Turing test, which is also summarized below as a *Qualitative Sugeno Model* [44] defined by the following two fuzzy if-then (Sugeno, [53]) rules written in Cognitive Dialect [37], as in [44]:

i)   $(!:) \{\{(!:)[ t \rightarrow (C \in IMP_H) ]\} \leftrightarrow \{(!:)[ t \rightarrow (d'(C) = 0) ]\}\}$,

ii)  $(!:) \{\{(!:)[ t \rightarrow (C \in GEN_H) ]\} \leftrightarrow \{(!:)[ t \rightarrow (d'(C) = 1) ]\}\}$,

where all symbols have the above defined meanings excepting d' which is an *ad hoc* unspecified decision function specific to the human agent.

Fig. 5.a and Fig. 5.b taken together describe a partially supervised iris recognition system in which the human agent (the supervisor) partially corrects the automated artificial understanding of iris recognition by taking the correct biometric decision for all iris pairs within the set of pairs artificially detected and labeled as undecidable ($UND_A$). The left-side dashed arrow marks these corrections in the figure showing that the human agent decides correctly even for the iris pairs whose similarity scores are irrelevant to the artificial agent. Unlike the errors associated to the pairs within the set $UND_A$, which are visible and correctable, the errors that the artificial agent could make when forming the sets $IMP_A$ and $GEN_A$ are not (they are impossible to correct without exhaustive supervision, but exhaustive supervision is also impractical). The right-side dashed arrows mark these insurmountable and hidden errors in the figure.

When compared to the actual set of imposter and genuine pairs ($IMP_H$ and $GEN_H$), even during the exhaustive iris recognition test the sets $IMP_A$ and $GEN_A$ are correctly determined, the pessimistic predictions assumed when defining the F3VDM (Fig. 4) show that in the future exploitation of the supervised iris recognition system (Fig. 5.a and Fig. 5.b) the biometric decisions are expected to be *almost correct* accordingly to the values $v_1$ and $v_2$ which both define the F3VDM. This is why the two calibration values $v_1$ and $v_2$ must be chosen as low as possible.

Of course, such a detailed investigation on the recognition errors is not possible when the exhaustive test of iris recognition produces results that are incompatible with designing a performant F3VDM. For example, if the overlap between the experimental genuine and imposter score distributions stretches between their means, the recognition results are incompatible with designing a performant F3VDM defined by imposing high safety conditions. Smaller the overlap, greater the chances to define a performant F3VDM satisfying higher safety conditions and supporting a wider safety band between the two distribution of scores.

## 6. Experimental results

This section presents the results of six exhaustive iris recognition tests undertaken for the database [52], the comparison to the results previously obtained in [38] on the same database and some insightful comments regarding the results and the comparison. Three of these tests assume the single eye enrollment scenario (each eye defines an identity), whereas the other three adopt the dual iris approach proposed in [45] and [46] in which the digital identity is defined using both eyes of an individual.

We consider that all experimentally results presented here are especially relevant in the context of the newly proposed AFKD (Automatic Formal Knowledge Discovery) technique for iris recognition defined in [44] as being an informed or uninformed search within the meta-theory of iris recognition whose goal is to identify a better iris recognition theory (and a better practice of iris recognition) based on genetic mutations, on logical and intelligent evolution. Nothing prevents us from considering that the paper [38] - for example, presents a formal theory of iris recognition in which all processing steps (from iris segmentation to the computation of the similarity score) are the genes of that formal theory of iris recognition, the genes of an individual ('point') belonging in the meta-theory of iris recognition which, on its turn, it is a virtual population space of virtually possible individuals that the evolution process could produce. In this context, we analyze how much stability the iris recognition results presented in [38] can prove when a simulated evolution process slightly changes the genes of the given formal theory of iris recognition. The first three tests assume changes in the pupil segmentation procedure (initially proposed in [36] and later reused in [38]) by adding a correction to the pupil radius as specified in the Section 4 and also adopt changes when the Haar Hilbert encoders used in [38] are replaced by the simplest Haar-Hilbert encoder described above (Section 4.2) or by the Combined Log-Gabor & Haar-Hilbert encoder (see Section 4.3). Another kind of mutation is simulated within the last three tests of iris recognition in which the similarity score fuses the information from four channels (see Fig. 3 and the subsequent comments in Section 4.3), two channels for each eye of an individual.

The experimental results presented along this section indicate that all of these simulated mutations are possible steps within a natural evolution process able to minimize the recognition errors in terms of EER, OEE or POEE. However, detecting these logical, intelligent, adequate and *meaningful* changes automatically, without the human supervision is an open problem. In our view, it is a huge difference between what we would call *a meaningful search* and a randomized search which is very often meaningless (just a lucky guess) even when it is effective. On the other hand, in our case, regardless the procedure by which it is discovered, *a meaningful mutation causes the decrease of iris recognition error*. Therefore, it is clear for us that the above stated open problem is a small facet of the bigger problem of understanding and formalizing the causality, which on its turn is considered by many (starting with Zadeh) to be another important and open problem.

Summarizing, in the context of AFKD techniques introduced in [44], the iris recognition tests undertaken here just exemplify the concept of *meaningful mutations* for the formal theory of iris recognition presented in [38] and rise a question on the open problem of their automated and unsupervised detection, which is classified here as a problem of causal structure/relation automated and unsupervised discovery / learning.

### 6.1 Experimental results obtained for the single eye enrollment scenario

Table 1 and Fig. 6 present the results of three exhaustive iris recognition tests (T1, T2 and T3) undertaken in the single eye enrollment scenario on the database [52]. The tests are labeled according to the encoder that they make use of: Haar-Hilbert (HH), Log-Gabor (LG) and combined Haar-Hilbert & Log-Gabor (HH&LG). Each time when a test uses HH encoder, Table 1 displays the corresponding size of the Hilbert filter.

The upper half of Table 1 shows the statistics of imposter and genuine scores (in terms of mean, standard deviation, degrees of freedom) obtained in each test and also the values corresponding to the evaluation criteria such as decidability index (d'), Fisher ratio(FR), overlap ($O_1$), the value of Equal Error Rate (EER), Maximum Imposter Score (MIS), minimum Genuine Score (mGS), the False Reject Rates (FRR) and the False Accept Rates (FAR) at MIS and mGS.

The bottom half of Table 1 illustrates (in terms of recognition threshold, FAR, FRR, OFA and OFR values) the behavior of the biometric system considered in some functioning regimes defined by imposing certain ranges for the FAR (1E-3, 1E-4, 1E-5) and FRR (2E-2, 1E-2, 1E-3) values.

Daugman said in [5] that combining two biometric tests, a weaker and a stronger one, the result could be a test with an average performance that "*will lie somewhere between that of the two tests conducted individually*" [5], but he also said that there are cases in which the recognition results can improve. Such a case is observed here in the third column of Table 1 for the test that makes use of combined HH&LG encoder. The statistics of the two distributions of scores, the overlap, the EER value and FAR(mGS), all of them certify an improvement in the iris recognition performance.

When compared to the results presented in Fig. 1, Fig. 2 and Fig. 3 from [38], the results of exhaustive iris recognition tests T1, T2, T3 illustrated here in Table 1, and Fig. 6 show an important improvement in the iris recognition performance. Since the test T1 from here produces 1Kb iris codes, it is comparable to the tests T1-T6 from [38] that produce 1Kb or even 4Kb iris codes. There are two easy ways to notice the differences between the recognition tests

**Table 1 Three exhaustive iris recognition tests on [52] in single eye enrollment scenario**

| Encoder: | Haar-Hilbert (T1) | Log-Gabor (T2) | HH&LG encoder (T3) |
|---|---|---|---|
| System parameters: | | | |
|    Iris code size | 8x128 | 16x256 | 8x128, 16x256 |
|    Hilbert filter size | 8 | Not applicable | 8 |
| Inter-class distribution: | | | |
|    Mean/Standard deviation | 0.5060 / 0.0222 | 0.5024 / 0.0196 | 0.5041 / 0.0187 |
|    Degrees-of-freedom | 508 | 650 | 712 |
| Intra-class distribution: | | | |
|    Mean/Standard deviation | 0.7804 / 0.0564 | 0.7748 / 0.0598 | 0.7775 / 0.0574 |
|    Degrees-of-freedom | 54 | 49 | 52 |
| Evaluation criteria: | | | |
|    Decidability / Fisher's ratio | 6.4022 / 20.494 | 6.1186 / 18.7187 | 6.4004 / 20.4824 |
|    **Overlap / EER** | **3.8085E-2 / 2.7836E-4** | **5.2734E-2 / 1.4021E-4** | **2.3250E-2 / 8.8879E-5** |
|    MIS / mGS | 0.6143 / 0.5761 | 0.5967 / 0.5439 | 0.5892 / 0.5659 |
|    FAR(MIS) / FRR(MIS) | 2.0531E-6 / 1.9340E-2 | 2.0531E-6 / 1.9367E-2 | 2.0531E-6 / 1.9381E-2 |
|    FAR(mGS) / FRR(mGS) | 1.1045E-3 / 1.0589E-4 | 1.0795E-2 / 1.0589E-4 | 4.9890E-4 / 1.0589E-4 |
| FUNCTIONING REGIMES | | | |
| FRR near 0.02: | | | |
|    threshold (t) / FRR(t) | 0.6602 / 1.9485E-2 | 0.64568 / 1.9379E-2 | 0.65821 / 1.9168E-2 |
|    FAR(t) / POFA(t) | 0 / 8.7549E-12 | 0 / 8.8696E-13 | 0 / 8.8818E-16 |
| **FRR near 0.01:** | | | |
|    threshold (t) / FRR(t) | 0.64726 / 9.9545E-3 | 0.63168 / 9.3668E-3 | 0.64221 / 9.8486E-3 |
|    FAR(t) / POFA(t) | **0 / 4.1368E-10** | **0 / 1.169E-10** | **0 / 6.028E-13** |
| FRR near 1E-3: | | | |
|    threshold (t) / FRR(t) | 0.60126 / 9.5309E-4 | 0.59368 / 9.5309E-4 | 0.59721 / 9.5306E-4 |
|    FAR(t) / POFA(t) | 2.4637E-5 / 3.9319E-1 | 4.1062E-6 / 4.5138E-1 | 0 / 4.1306E-1 |
| FAR near 1E-3: | | | |
|    threshold (t) / FRR(t) | 0.57626 / **1.059E-4** | 0.56268 / **1.059E-4** | 0.56271 / **0** |
|    FAR(t) / POFA(t) | **9.7523E-4** / 1.6406E-3 | **1.0409E-3** / 2.409E-3 | **0.9547E-3** / 2.1117E-3 |
| FAR near 1E-4: | | | |
|    threshold (t) / FRR(t) | 0.59226 / **5.2942E-4** | 0.57618 / **2.118E-4** | 0.57471 / **1.059E-4** |
|    FAR(t) / POFA(t) | **1.0676E-4** / 1.2556E-4 | **9.4444E-5** / 2.2631E-4 | **9.6497E-5** / 2.3129E-4 |
| FAR near 1E-5: | | | |
|    threshold (t) / FRR(t) | 0.60776 / 1.9062E-3 | 0.59268 / 8.4719E-4 | 0.58471 / 3.177E-4 |
|    FAR(t) / POFA(t) | 8.2125E-6 / 6.5154E-6 | 1.0266E-5 / 6.8479E-6 | 8.2125E-6 / 2.7262E-5 |

T1-T6 from [38] and the test T1 from here:

The first one is to compare the evaluation criteria (especially the EER) and the functioning regimes near a FRR of 1E-2 presented in the Table 1 from [38] for the former tests to the similar values computed for the latter test and presented in Table 1 from here. For example, the EER in the latter test is better (smaller) than all EER values of the former tests (T1-T6 from [38]), despite the fact that some of these tests (namely the tests T1-T3 from [38]) use 4Kb binary iris codes.

The second one is to compare the ROC curves obtained for the tests T1-T3 from [38] (see Fig. 3.a in [38]) to those obtained for the tests T1-T3 from here and presented in Fig. 6.e. Despite the former tests use 4Kb iris codes whereas the latter tests use 1Kb iris codes, the ROC curves for the former tests (see Fig. 3.a in [38]) are weaker than those of the latter tests (Fig. 6.e): the best ROC curve in Fig.3.a from [38] (which starts its descent from a value close to 3E-3) is weaker than the weakest ROC curve in Fig. 6.e (which starts its descent from a value close to 2E-3), and the best EER value in Fig. 3.a from [38] (5.4397E-4) is also weaker than the weakest EER value in Fig. 6.e (2.7836E-4). Hence, the simulated mutations discussed in the beginning of Section 6 are indeed *meaningful* whereas the evolution simulated through these mutations is a *natural evolution* that changes the genes of the iris recognition theory illustrated in [38] toward the error minimization.

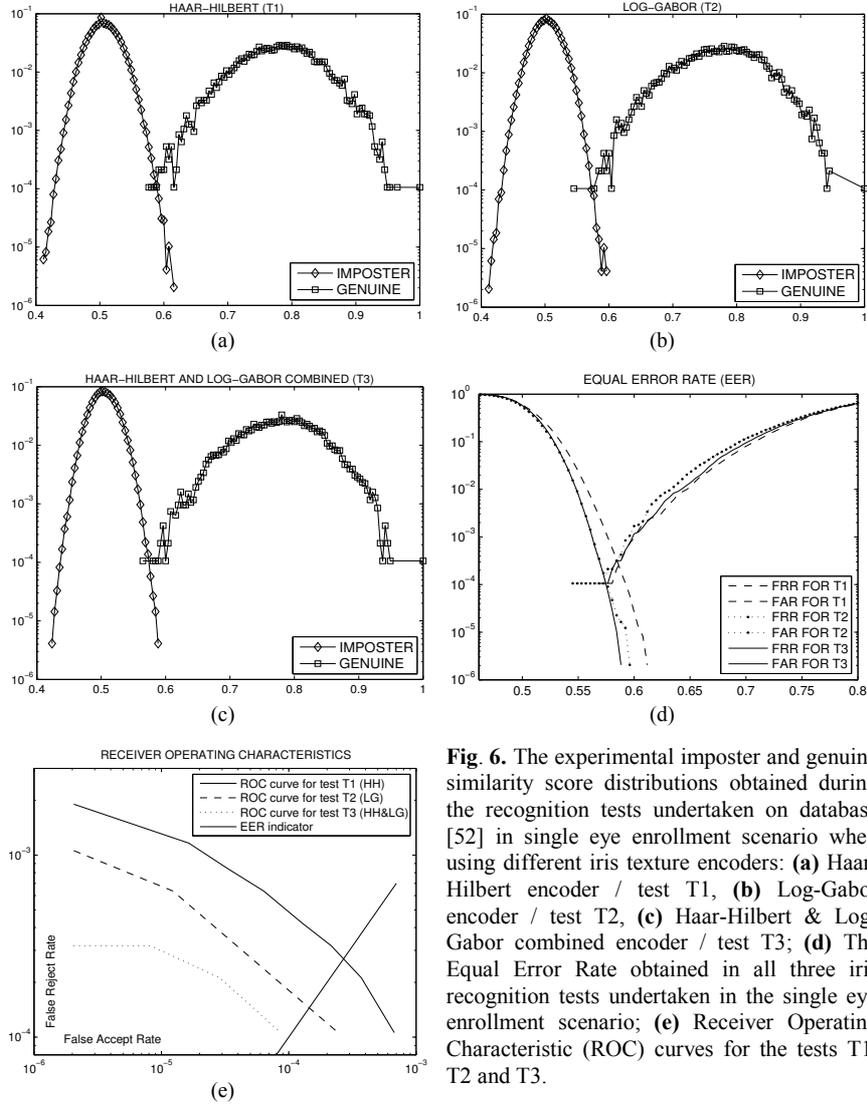

**Fig. 6.** The experimental imposter and genuine similarity score distributions obtained during the recognition tests undertaken on database [52] in single eye enrollment scenario when using different iris texture encoders: **(a)** Haar-Hilbert encoder / test T1, **(b)** Log-Gabor encoder / test T2, **(c)** Haar-Hilbert & Log-Gabor combined encoder / test T3; **(d)** The Equal Error Rate obtained in all three iris recognition tests undertaken in the single eye enrollment scenario; **(e)** Receiver Operating Characteristic (ROC) curves for the tests T1, T2 and T3.

## 6.2. Experimental results obtained for the dual iris approach

Table 2 and Fig. 7 present the results of three exhaustive iris recognition tests (T4, T5 and T6) undertaken in the dual iris enrollment scenario (dual iris approach [45], [46]) on the database [52]. The tests are labeled according to the encoder that they make use of: Haar-Hilbert (HH), Log-Gabor (LG) and combined Haar-Hilbert & Log-Gabor (HH&LG). Each time when a test uses HH encoder, Table 2 displays the corresponding size of the Hilbert filter. These three iris recognition tests assume that each person is enrolled with both eyes and any recognition request is treated using the both eye of the candidate.

The upper half of Table 2 shows the statistics of imposter and genuine scores (in terms of mean, standard deviation, degrees of freedom) obtained in each test and also the values corresponding to the evaluation criteria such as decidability index (d'), Fisher ratio(FR), overlap, the value of Equal Error Rate (EER), Maximum Imposter Score (MIS), minimum Genuine Score (mGS), the Pessimistic Odds of False Reject (POFR) and of the False Accept (POFA) at MIS and mGS.

The bottom half of Table 2 illustrates (in terms of recognition threshold, FAR, FRR, POFA and POFR values) the behavior of the biometric system considered in some functioning regimes defined by imposing certain ranges for the pessimistic odds POFA and POFR (2E-2, 1E-2, 1E-3, 1E-4, 1E-5, 1E-6).

The statistics of the two distributions of scores obtained for the sixth test (last column in Table 2), but especially the overlap, EER, POEE, POFR(MIS) and POFA(mGS) values, all of them confirm the improvement in the iris recognition performance.

Comparing the results obtained for the single eye enrollment scenario (T1-T3) to those obtained for the dual iris approach (T4-T6) is easy if we look at the data

**Table 2 The differences between the two encoders and the combination between them**

| Encoder | Haar-Hilbert (T4) | Log-Gabor (T5) | HH&LG encoder (T6) |
|---|---|---|---|
| System parameters: | | | |
| Iris code size | 8x128 | 16x256 | 8x128, 16x256 |
| Hilbert filter size | 8 | - | 8 |
| Inter-class distribution: | | | |
| Mean / Standard deviation | 0.5118 / 0.0161 | 0.5039 / 0.0144 | 0.5078 / 0.0138 |
| Degrees-of-freedom | 958 | 1211 | 1307 |
| Intra-class distribution: | | | |
| Mean / Standard deviation | 0.7795 / 0.0462 | 0.7737 / 0.0491 | 0.7766 / 0.0472 |
| Degrees-of-freedom | 80 | 73 | 78 |
| Evaluation criteria: | | | |
| Decidability index | 7.7312 | 7.4585 | 7.7205 |
| Fisher's ratio | 29.8857 | 27.8148 | 29.8031 |
| Overlap | -3.6935E-2 | -5.0329E-2 | -6.1464E-2 |
| **EER / POEE** | **0 / 6.2979E-5** | **0 / 7.9905E-5** | **0 / 5.1609E-5** |
| MIS / mGS | 0.5843 / 0.6212 | 0.5581 / 0.6083 | 0.5645 / 0.6259 |
| **POFA(mGS) / POFR(MIS)** | **8.0974E-10 / 9.2260E-5** | **1.4427E-11 / 5.5616E-5** | **1.2357E-11 / 3.5077E-5** |

Table 2 (continued)

| Encoder | Haar-Hilbert (T4) | Log-Gabor (T5) | HH&LG encoder (T6) |
|---|---|---|---|
| **FUNCTIONING REGIMES** | | | |
| POFA near 2E-2: | | | |
| threshold (t) | 0.545 | 0.540 | 0.540 |
| FRR(t) / POFR(t) | 0 / 3.4197E-6 | 0 / 1.2097E-5 | 0 / 4.4961E-6 |
| FAR(t) / POFA(t) | 2.4258E-2 / 2.5854E-2 | 1.3450E-2 / 1.9864E-2 | 1.1866E-2 / 1.3561E-2 |
| POFA near 1E-2: | | | |
| threshold (t) | 0.550 | 0.540 | 0.540 |
| FRR(t) / POFR(t) | 0 / 5.3208E-6 | 0 / 1.2097E-5 | 0 / 4.4961E-6 |
| FAR(t) / POFA(t) | 1.2866E-2 / 1.3277E-2 | 4.5916E-3 / 9.0645E-3 | 1.1866E-2 / 1.3561E-2 |
| **POFA near 1E-3:** | | | |
| threshold (t) | 0.565 | 0.550 | 0.555 |
| FRR(t) / POFR(t) | **0 / 1.5028E-5** | **0 / 1.6337E-5** | **0 / 1.5808E-5** |
| FAR(t) / POFA(t) | 9.4166E-4 / **1.2013E-3** | 2.4166E-4 / **1.4610E-3** | 3.5E-4 / **7.7472E-4** |
| **POFA near 1E-4:** | | | |
| threshold (t) | 0.580 | 0.565 | 0.565 |
| FRR(t) / POFR(t) | **0 / 6.2979E-5** | **0 / 7.9905E-5** | **0 / 3.5077E-5** |
| FAR(t) / POFA(t) | 8.3333E-6 / **5.8429E-5** | 0 / **4.9093E-5** | 0 / **7.1695E-5** |
| POFA near 1E-5: | | | |
| threshold (t) | 0.590 | 0.570 | 0.575 |
| FRR(t) / POFR(t) | 0 / 1.34E-4 | 0 / 1.1393E-4 | 0 / 7.5313E-5 |
| FAR(t) / POFA(t) | 0 / 5.4750E-6 | 0 / 1.3262E-5 | 0 / 4.5147E-6 |
| POFA near 1E-6: | | | |
| threshold (t) | 0.595 | 0.580 | 0.580 |
| FRR(t) / POFR(t) | 0 / 1.9298E-4 | 0 / 2.2644E-4 | 0 / 1.09E-4 |
| FAR(t) / POFA(t) | 0 / 1.5065E-6 | 0 / 7.3899E-7 | 0 / 9.7924E-7 |
| POFR near 0.02: | | | |
| threshold (t) | 0.675 | 0.660 | 0.670 |
| FRR(t) / POFR(t) | 1.5789E-2 / 2.1133E-2 | 1.3684E-2 / 1.8893E-2 | 1.7052E-2 / 2.1403E-2 |
| FAR(t) / POFA(t) | 0 / 8.7330E-20 | 0 / 1.3005E-22 | 0 / 4.9003E-26 |
| **POFR near 1E-2:** | | | |
| threshold (t) | 0.660 | 0.645 | 0.655 |
| FRR(t) / POFR(t) | 4.6315E-3 / 1.0289E-2 | 5.2631E-3 / 9.5060E-3 | 4.6315E-3 / 1.0605E-2 |
| FAR(t) / POFA(t) | 0 / **1.0962E-16** | 0 / **7.0675E-19** | 0 / **7.6187E-22** |
| **POFR  near 1E-3:** | | | |
| threshold (t) | 0.620 | 0.605 | 0.610 |
| FRR(t) / POFR(t) | 0 / 1.0527E-3 | 0 / 1.1055E-3 | 0 / 8.4480E-4 |
| FAR(t) / POFA(t) | 0 / **8.0974E-10** | 0 / **1.1051E-10** | 0 / **1.2945E-11** |
| POFR near 1E-4: | | | |
| threshold (t) | 0.585 | 0.570 | 0.580 |
| FRR(t) / POFR(t) | 0 / 9.2260E-5 | 0 / 1.1393E-4 | 0 / 1.09E-4 |
| FAR(t) / POFA(t) | 0 / 1.8530E-5 | 0 / 1.3262E-5 | 0 / 9.7924E-7 |
| POFR near 1E-5: | | | |
| threshold (t) | 0.555 | 0.535 | 0.550 |
| FRR(t) / POFR(t) | 0 / 8.2073E-6 | 0 / 8.1053E-6 | 0 / 1.0482E-5 |
| FAR(t) / POFA(t) | 6.05E-3 / 6.3790E-3 | 1.3450E-2 / 1.9864E-2 | 1.2166E-3 / 2.2083E-3 |
| POFR near 1E-6: | | | |
| threshold (t) | 0.530 | 0.510 | 0.525 |
| FRR(t) / POFR(t) | 0 / 8.6194E-7 | 0 / 9.7548E-7 | 0 / 1.1868E-6 |
| FAR(t) / POFA(t) | 1.2725E-2 / 1.2961E-2 | 3.4185E-1 / 3E-1 | 1.0679E-1 / 1.0489E-1 |

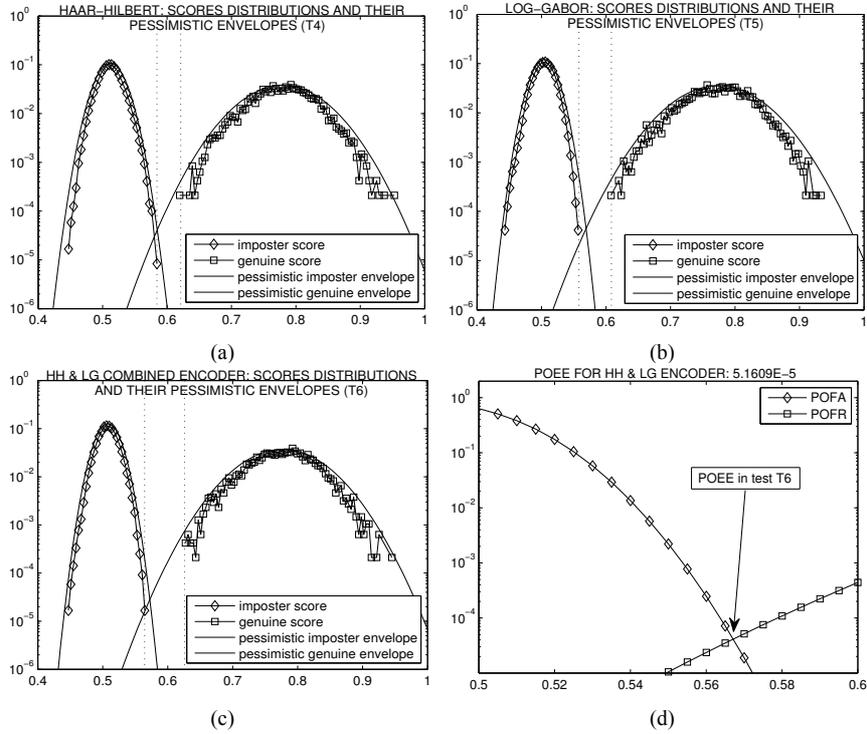

**Fig. 7.** The genuine and imposter score distributions and their pessimistic approximations for each iris recognition test undertaken on [52] in the dual iris approach ([45], [46]) when using: **(a)** Haar-Hilbert, **(b)** Log-Gabor, **(c)** Combined Haar-Hilbert & Log-Gabor iris texture encoders. **(d)** - The Pessimistic Odds of Equal Error (POEE) obtained for the third iris recognition tests presented in Table 2 (dual iris approach, combined HH&LG iris encoder).

displayed in Table 1 and Table 2 for the functioning regimes in which the value of POFR is near 1E-2 or 1E-3 and also for the functioning regimes in which the value of POFA is near 1E-3 or 1E-4.

The values POFA(mGS) / POFR(MIS) within Table 2 allow us to define very safe Fuzzy 3-Valent Disambiguated Models (F3VDM-s) [42] of iris recognition in which the safety band is determined by MIS and mGS. Alternatively, one could define the extremities of the safety band as preimages of two imposed values for POFA and POFR. Fig. 8.a and Fig. 8.b illustrate such F3VDM-s obtained for the dual iris approach by imposing pairs of security restrictions. Another way to define a F3VDM is to target a certain value for POFA (limiting the odds of false accepts) and a certain discomfort rate cumulated over the safety interval – see Fig. 8.c.

From an experimental point of view, the importance of defining F3VDM-s for the exhaustive iris recognition tests T1-T6 resides in selecting those pairs of irides for which the artificial understanding of their similarity or non-similarity is the weakest, i.e. the pairs of irides that support and prove the fuzzy separation between the artificially perceived concepts of *genuine* and *imposter* comparisons. The defuzzification is achieved by classifying these pairs as artificially undecidable (see Fig. 5.a and Fig. 5.b) and by producing a controlled loss in user's comfort quantifiable through the total discomfort rate cumulated over the safety interval (as described in Fig. 8.a, Fig. 8.b and Fig. 8.c).

The pairs labeled as being artificially undecidable could be forwarded to the human agent for analyzing and verifying the quality of iris recognition during a Turing test. Another possibility is to use these problematic pairs for training discriminant and witness directions ([43], [44]) as robust digital identities or to decide the similarity of their components using previously trained discriminant

<br>

**(a)**

| I | [0.6050, 1] | Genuine pairs | False Accept Rate: POFA(0.6050) = 1.0777E-10 <br> True Accept Safety: 1-POFA(0.6050) | |
|---|---|---|---|---|
| O | (0.5800, 0.6050) | Artificially Undecidable Pairs | INCERTITUDE — Genuine Discomfort Rate < POFR(0.6050) ≈ 6.1282E-4; <br> Imposter Discomfort Rate < POFA(0.5800) ≈ 9.7924E-7. <br> ──────────────── <br> Total Discomfort Rate ≈ **6.1379E-4** | DISCOMFORT / SECURITY |
| D | [0, 0.5800] | Imposter pairs | False Reject Rate: POFR(0.5800) = 1.09E-4 <br> False Reject Safety: 1-POFR(0.5800) | |

<br>

**(b)**

| I | [0.6050, 1] | Genuine pairs | False Accept Rate: POFA(0.6050) = 1.0777E-10 <br> True Accept Safety: 1-POFA(0.6050) | |
|---|---|---|---|---|
| O | (0.5500, 0.6050) | Artificially Undecidable Pairs | INCERTITUDE — Genuine Discomfort Rate < POFR(0.6050) ≈ 6.1282E-4; <br> Imposter Discomfort Rate < POFA(0.5500) ≈ 2.2083E-3. <br> ──────────────── <br> Total Discomfort Rate ≈ **2.8211E-3** | DISCOMFORT / SECURITY |
| D | [0, 0.5500] | Imposter pairs | False Reject Rate: POFR(0.5500) = 1.0482E-5 <br> False Reject Safety: 1-POFR(0.5500) | |

<br>

**(c)**

| I | [0.6050, 1] | Genuine Pairs | False Accept Rate: POFA(0.6050) = 1.0777E-10 <br> True Accept Safety: 1-POFA(0.6050) | |
|---|---|---|---|---|
| O | (0.5450, 0.6050) | Artificially Undecidable Pairs | INCERTITUDE — Genuine Discomfort Rate < POFR(0.6050) ≈ 6.1282E-4; <br> Imposter Discomfort Rate < POFA(0.5450) ≈ 9.3871E-3. <br> ──────────────── <br> Total Discomfort Rate ≈ **1E-2** | DISCOMFORT / SECURITY |
| D | [0, 0.5450] | Imposter Pairs | False Reject Rate: POFR(0.5450) = 6.8936E-6 <br> False Reject Safety: 1-POFR(0.5450) | |

**Fig. 8**. Three fuzzy 3-valent disambiguated decisional models (F3VDM) obtained for the dual iris approach by imposing the following pairs of security restrictions: **(a)** r = 0.58 ≈ POFR$^{-1}$(1E-5) and a = 0.6050 ≈ POFA$^{-1}$(1E-10), **(b)** r = 0.55 ≈ POFR$^{-1}$(1E-5) and a = 0.6050 ≈ POFA$^{-1}$(1E-10), **(c)** a = 0.6050 ≈ POFA$^{-1}$(1E-10) and r = 0.5450 is determined such that the total discomfort rate on the safety interval (r,a) to be near 1E-2.

and witness directions (which belong to the category of soft-biometric memories / information, [2]). Any such pair of irides corresponds to a pair formed with the identity of the current candidate (representing the individual who claims an enrolled identity) and with the claimed enrolled identity, a pair of identities for which the comparison between the corresponding irides is not relevant enough for predicting the relation between the two identities, a pair of identities whose relation can not be decided accurately using a single-biometric system based on iris recognition. This explains why we consider that the undecidable pairs are perfect candidates for combining hard- and soft-biometric information [2] or for fusing single-biometrics into cascaded multi-biometrics systems [50] based on iris recognition, palm-vein [20], face [21], fingerprint [26] or ear [48] recognition. All of these came into our attention for future multi-biometrics joint studies and works. We do not exclude the possibility of simulating a parallel multi-biometrics system in our future works, but the cascaded multi-biometrics architecture [50] looks more promising now.

## 7. Instead of conclusion - F3VDM-s for the dual iris approaches: safety vs. comfort, visible vs. hidden recognition errors

In any single-biometric system based on bimodal iris recognition, any recognition threshold t determines a statistically predicted level of comfort associated to the honest users that claim their actual identities - encoded as FRR(t), OFR(t) or POFR(t), and a statistically predicted level of safety associated to the honest users that reject identities that are associated to other users – safety level encoded as FAR(t), OFA(t) or POFA(t). This fact occurs regardless if the system is based on single-eye enrollment scenario or on the dual eyes enrollment scenario (dual iris approach, [45], [46]), but because the recognition results are better in the latter case, we choose to continue our investigation for this case only. However, the safety requirements for a biometric system could be other than the pairs (POFA(t), POFR(t)). Fig. 8 presents such cases in which the safety restrictions define the safety interval instead of a recognition threshold.

Let us imagine a biometric world-wide network like that presented in [41] in which the candidate iris code CIC is extracted on a biometric terminal and carried to a central server through a safe communication protocol, without any other additional data. Hypothetically, the central server practice the bimodal (statistical) iris recognition in the dual iris approach described above in this chapter (Table 2, Fig. 7 and Fig. 8) and it is supposed to be able to classify the candidate iris code CIC to the appropriate enrolled identity or to infer that the candidate iris code CIC represents for sure an identity which has not been enrolled yet in the biometric network. Instead any of these, let us suppose that the similarity score measuring the proximity of the candidate iris code CIC to a certain enrolled identity belongs to the safety band (or even matches the threshold corresponding to the POEE value). What is the decision that the server should take? Is there a good decision that the system could take? It is easy to verify that any decision the system could take, other than requesting a new iris code sample from the same candidate, is associated with a minimum (null) level of confidence, hence by doing otherwise, the system would deliver the correct decision only by pure chance (by flipping a coin). Of course, if we choose a recognition threshold and allow the system to classify the current claim accordingly, still the level of confidence of any biometric decision that system would take in this case stays (fuzzy) null, despite the arbitrary position of the chosen threshold. The case described above is a case of recognition error. Narrowing the safety band of a F3VDM up to a recognition threshold could hide such recognition errors from the view of inexperienced eye, but will never change the nature of these errors and will never prevent them for happening. When these errors are hidden, the safety of the system is (in fact) low and user's comfort is high. The importance of the Fuzzy 3-Valent Disambiguated Models of iris recognition (recently proposed in [44] and [42]) is that they unveil a certain proportion of recognition errors, accordingly to the imposed safety restrictions specified as in Fig. 5 and Fig. 8 by two thresholds $t_1 = POFR^{-1}(v_1)$ and $t_2 = POFA^{-1}(v_2)$ which together define the safety interval. The recognition errors unveiled by a F3VDM could be corrected in a logical consistent manner by forwarding the appropriate cases to a human agent, to an artificial intelligent agent using soft-biometric data or even to the next level of a cascaded hard or soft multi-biometrics recognition system. However, the recognition errors hidden outside the safety interval will stay hidden indefinitely.

### Acknowledgments

We wish to acknowledge and thank N. Popescu-Bodorin for sharing with us his experience and an important part of the results that he has previously obtained during his PhD study [44], for kindly allowing us to continue his work, as presented above in this chapter, with a minimal effort.